\documentclass[sigconf]{acmart}

\usepackage{graphicx}
\usepackage{subcaption}
\usepackage{xcolor}
\usepackage{colortbl}

\AtBeginDocument{%
  }

\setcopyright{acmlicensed}
\copyrightyear{2018}
\acmYear{2018}
\acmDOI{XXXXXXX.XXXXXXX}
\acmConference[Conference acronym 'XX]{Make sure to enter the correct
  conference title from your rights confirmation email}{June 03--05,
  2018}{Woodstock, NY}
\acmISBN{978-1-4503-XXXX-X/2018/06}

\begin{document}

\title{CamWorldQA: Perceptual Quality Assessment of Camera-Controlled World Video Generation}

\author{Yunhe Li, Likun Wu, Sijing Wu, Xinyu Tian, Huiyu Duan, Yixuan Gao, Yunhao Li, Guangtao Zhai}
\author{Shanghai Jiao Tong University, Eindhoven University of Technology}

\renewcommand{\shortauthors}{Yunhe Li et al.}

\begin{abstract}
Recent advances in generative video models have enabled camera-controlled world video generation, allowing models to synthesize videos under user-defined camera trajectories. However, existing video quality assessment (VQA) methods are mainly developed for natural videos and fail to capture the unique perceptual characteristics of camera-controlled generation, such as viewpoint consistency, motion coherence, and content preservation. In this work, we introduce \textbf{CamWorldQA}, the first benchmark for perceptual quality assessment of camera-controlled world video generation. CamWorldQA contains 720 generated videos produced by 6 representative generation methods from 20 diverse source videos under 6 camera trajectories, where each video is annotated with a human-rated perceptual quality score through subjective experiments. Furthermore, we propose \textbf{CWQA}, a no-reference quality assessment network with three complementary branches that extract spatial features, temporal motion features and optical flow features to jointly predict quality scores. Extensive experiments demonstrate that CWQA achieves superior performance over existing quality assessment methods on the CamWorldQA dataset.
\end{abstract}



\begin{teaserfigure}
\includegraphics[width=\textwidth]{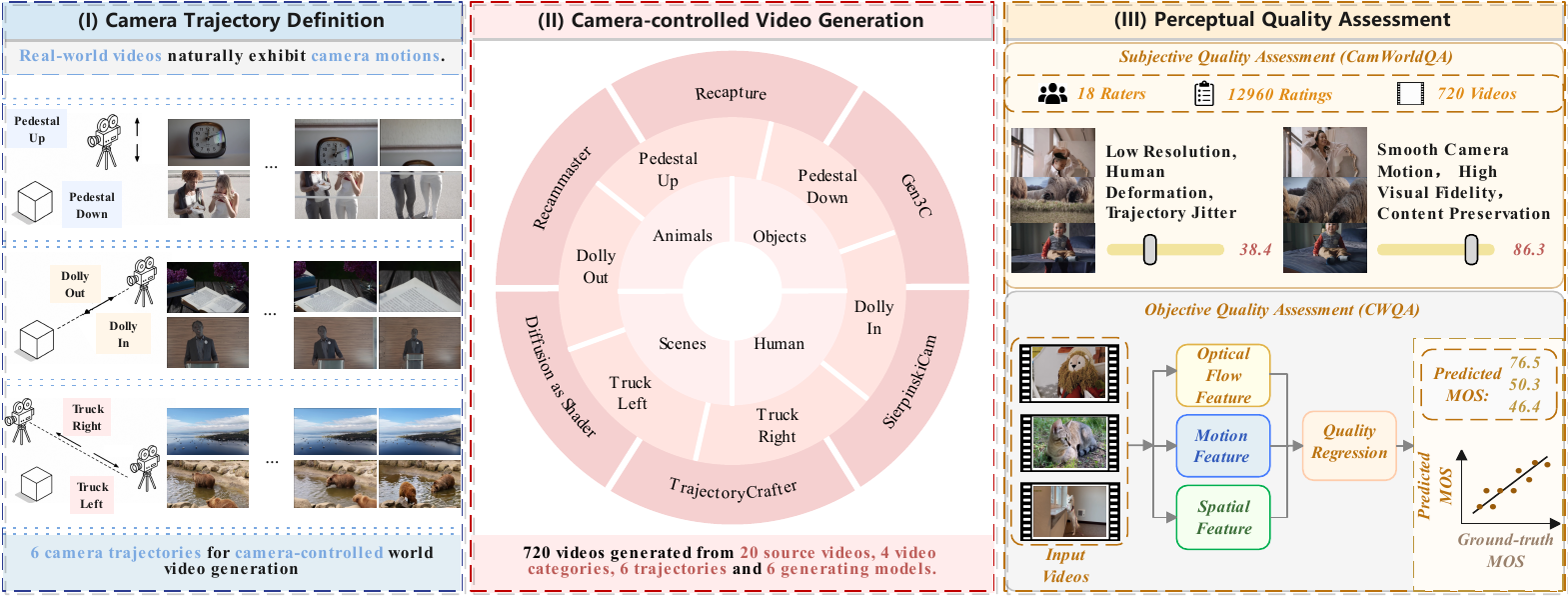}
\caption{Overview of camera-controlled world video quality assessment, including camera trajectory definition, CamWorldQA dataset construction, and the proposed CWQA model.}
\label{fig:teaser}
\end{teaserfigure}

\maketitle

\section{Introduction}

Recent advances in video generation have enabled camera-controlled video synthesis \cite{bai2025recammaster,yu2025trajectorycrafter,park2026redirector,ren2025gen3c}, which takes a source video and a new camera trajectory as inputs to generate a new video following the new trajectory. It has numerous applications including film production, virtual reality, embodied intelligence, and interactive world simulation. However, camera-controlled generation requires both visual realism and consistent viewpoint transitions. Current methods may still suffer from inaccurate trajectory following, viewpoint inconsistency, temporal flickering, and unintended content changes, making reliable perceptual quality assessment essential for evaluating and improving these methods.

Existing video quality assessment (VQA) methods have been primarily developed for natural or user-generated videos, focusing on distortions such as compression artifacts, blur, noise, and motion distortions \cite{wu2022fast, wu2023exploring}. Although recent approaches incorporate spatial-temporal representations and motion-aware features, they are not specifically designed to capture generation-specific quality factors in camera-controlled videos, including viewpoint inconsistency, incoherent camera motion, and unintended content deformation. Furthermore, the absence of a dedicated subjective benchmark limits the systematic evaluation and comparison of quality assessment methods for generated camera-controlled videos.

In light of these facts, we introduce CamWorldQA, a benchmark specifically designed for perceptual quality assessment of camera-controlled video generation. CamWorldQA contains 720 generated videos constructed from 20 diverse real-world source videos using 6 representative generation methods and 6 camera trajectories, covering 4 content categories and diverse generation qualities. We conduct a subjective study to collect human perceptual quality annotations, where participants evaluate the generated videos by considering viewpoint consistency, motion coherence and content preservation, with the mean opinion scores (MOSs) for quality assessment methods.

Based on this benchmark, we further propose CWQA, a no-reference quality assessment framework for camera-controlled generated videos. Specifically, we employ a spatial branch to extract multi-level appearance features and a temporal branch to capture high-level video dynamics. Moreover, an optical flow branch is introduced to explicitly characterize motion changes between consecutive frames. The three representations are projected into a common feature space and integrated through a lightweight branch-wise gating network that adaptively weights their contributions. Finally, the weighted features are concatenated and fed into a quality regression head to predict the perceptual quality score.


Our main contributions are summarized as follows:
\begin{itemize}
    \item We construct CamWorldQA, the first quality assessment dataset for camera-controlled video generation, including 720 generated videos from 20 source videos, 6 camera trajectories and 6 models with perceptual quality scores.
    
    \item Based on CamWorldQA, we systematically benchmark current camera-controlled video generation methods and representative video quality assessment methods and large multimodal models.
    
    \item We propose CWQA, a three-branch no-reference quality assessment framework that incorporates spatial, temporal and optical flow features with a branch-wise gating network to predict quality scores. The experimental results demonstrate the effectiveness of our method.
\end{itemize}

\section{Related Work}

\noindent\textbf{Camera-Controlled Video Generation.}
Recent advances in video generation based on diffusion have enabled increasingly
controllable synthesis through camera conditioning. Existing studies control camera motion using explicit poses,
predefined motion patterns, trajectory conditions, and geometric
representations~\cite{he2024cameractrl, wang2024motionctrl, xu2024camco,zheng2024cami2v,popov2025camctrl3d}.
 More recent works extend camera control to existing monocular videos, enabling video re-rendering and novel-view synthesis under specified camera trajectories while preserving the original scene content and dynamics~\cite{
bai2025recammaster,
yu2025trajectorycrafter,
park2026redirector,
zhang2025recapture,
van2024generative,
bian2025gs,
gu2025diffusion}. Other approaches incorporate geometric and 3D-aware priors, such as depth and reconstructed scene representations, to improve camera controllability and scene consistency~\cite{ren2025gen3c,wang2025cinemaster,hou2024training}. Despite these advances, camera-controlled videos can still suffer from inconsistent viewpoints, geometric deformation, temporal artifacts, and unintended content changes, motivating reliable perceptual quality assessment for this emerging type of generated video.

\noindent\textbf{Video Quality Assessment.}
Video quality assessment (VQA) aims to predict perceptual video quality in accordance with human judgments. Previous VQA studies have primarily focused on natural and user-generated content (UGC), covering both synthetic and authentic distortions~\cite{hosu2017konstanz,sinno2018large, li2026dhqa, wu2025fvq, wang2019youtube, ying2021patch}. Accordingly, no-reference VQA methods have evolved from handcrafted and hybrid quality representations~\cite{tu2021rapique,tu2021ugc,korhonen2019two} to deep spatial-temporal quality modeling~\cite{chen2021learning,sun2022deep, duan2025finevq}, with recent approaches further exploring efficient sampling, motion-aware representations, and complementary quality modeling~\cite{wu2022fast,kou2023stablevqa,wu2023exploring}. With the rapid development of generative models, perceptual quality assessment has recently been extended from UGC to AI-generated content (AIGC) \cite{ kou2024subjective, gao2025multi, wu2025hveval, li2025aghi}. Existing studies ~\cite{fang2026humanscore, wu2026multi, wu2025singinghead, zhang2025human, lu2024aigc} investigate generation-specific degradations from different perspectives, including visual fidelity, spatial-temporal quality, motion naturalness, aesthetics, and semantic consistency. However, camera-controlled video generation introduces unique quality challenges associated with viewpoint transitions, camera-induced motion, geometric consistency, and content preservation. Existing studies are not specifically designed to assess these camera-related degradations, motivating perceptual quality assessment tailored to camera-controlled video generation.

\begin{figure}[t]
\centering
\includegraphics[width=0.48\textwidth]{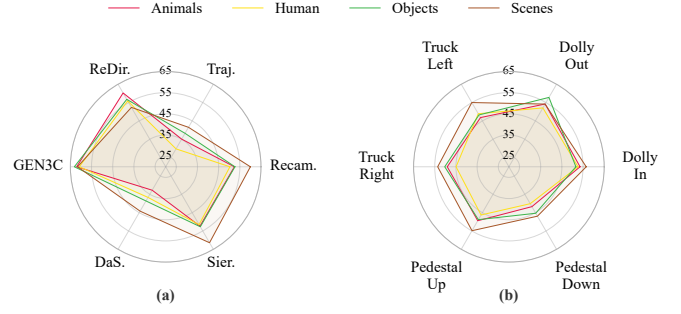}
\caption{
MOS variations across (a) generation methods and (b) camera trajectories.
}
\vspace{-3mm}
\label{fig:radar}
\end{figure}

\begin{figure*}[ht]
    \centering
    \includegraphics[width=\textwidth]{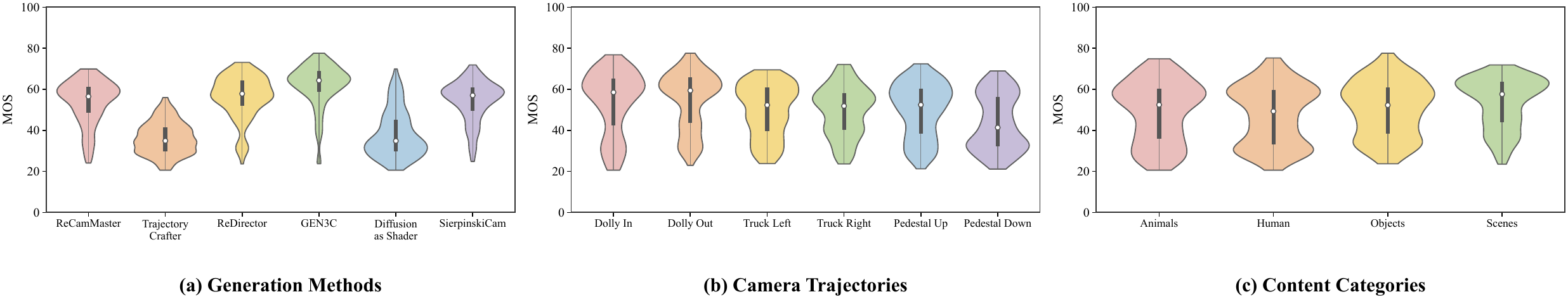}
    \caption{MOS distributions across (a) generation methods, (b) camera trajectories, and (c) video content categories.}
    \label{fig:model-dist}
\end{figure*}
\begin{figure}[ht]
    \centering
    \includegraphics[width=0.9\columnwidth]{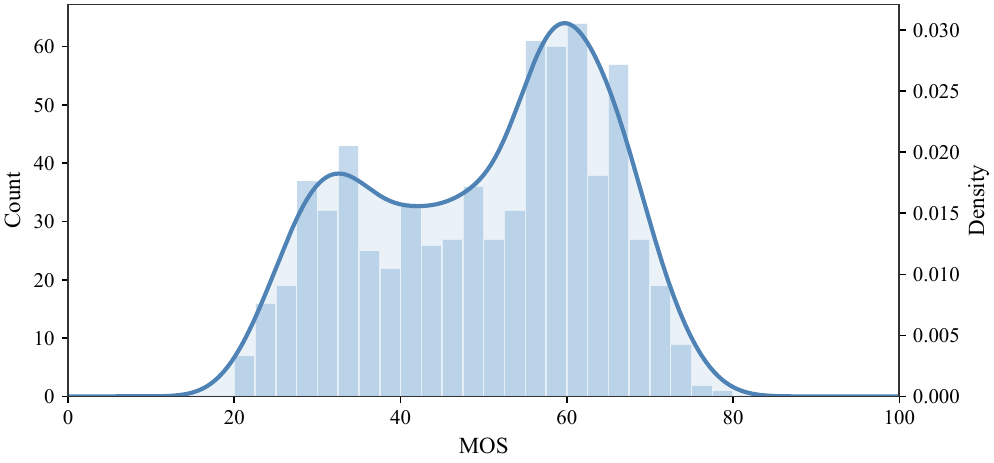}
    \vspace{-3mm}
    \caption{MOS distribution of CamWorldQA.}
    \label{fig:overall-dist}
\end{figure}

\section{Datasets}
\subsection{Data Preparation}

\noindent\textbf{Video source}: We collected 20 high-quality real-world videos with nearly static cameras. 
These videos are evenly divided into four categories: \textbf{Animals}, \textbf{Human}, \textbf{Objects}, and \textbf{Scenes}. 
Specifically, the \textbf{Animals} category includes domestic and wild animals. 
The \textbf{Human} category contains people in everyday activities. 
The \textbf{Objects} category covers both static and dynamic subjects. 
The \textbf{Scenes} category consists of natural and urban environments.

\noindent\textbf{Trajectories}: Based on frequency of use, we carefully designed six camera trajectories for generation: \textbf{dolly in}, \textbf{dolly out}, \textbf{truck left}, \textbf{truck right}, \textbf{pedestal up}, and \textbf{pedestal down}. We used trajectory-specific prompts to convert the six camera motions into the input format required by each generation model with AI assistants.

\subsection{Video Collection}
Based on these source videos and trajectories, we generated 720 camera-moving world
videos using six video-generating models, including ReCamMaster
\cite{bai2025recammaster}, TrajectoryCrafter \cite{yu2025trajectorycrafter}, ReDirector
\cite{park2026redirector}, GEN3C \cite{ren2025gen3c}, Diffusion as Shader
\cite{gu2025diffusion}, and SierpinskiCam \cite{wizadwongsa2026sierpinskicam}. We
utilized default settings and weights for these open-source models to generate
the videos. 
All the videos were preprocessed to a resolution of 832 × 480, a frame rate of 15 fps, and a total of 81 frames per clip before generation.

\begin{table}[t]
    \centering
    \caption{MOS statistics of videos generated by the six video generation methods in CamWorldQA.}
    \label{tab:mos-statistics}
    \setlength{\tabcolsep}{4pt}
    \resizebox{0.92\columnwidth}{!}{
    \begin{tabular}{lccccc}
        \toprule
        Method & Mean MOS $\uparrow$ & Std. & Min & Max & Rank \\
        \midrule
        GEN3C~\cite{ren2025gen3c}
        & 61.88 & 10.49 & 23.73 & 77.59 & 1 \\
        ReDirector~\cite{park2026redirector}
        & 56.31 & 10.59 & 23.63 & 73.09 & 2 \\
        SierpinskiCam~\cite{wizadwongsa2026sierpinskicam}
        & 54.49 & 9.92 & 24.83 & 71.86 & 3 \\
        ReCamMaster~\cite{bai2025recammaster}
        & 53.84 & 10.74 & 24.08 & 69.90 & 4 \\
        Diffusion as Shader~\cite{gu2025diffusion}
        & 38.07 & 11.40 & 20.59 & 69.90 & 5 \\
        TrajectoryCrafter~\cite{yu2025trajectorycrafter}
        & 36.15 & 7.87 & 20.58 & 55.99 & 6 \\
        \bottomrule
    \end{tabular}
    }
\end{table}

\begin{figure*}[ht]
    \centering
    \includegraphics[width=0.92\textwidth]{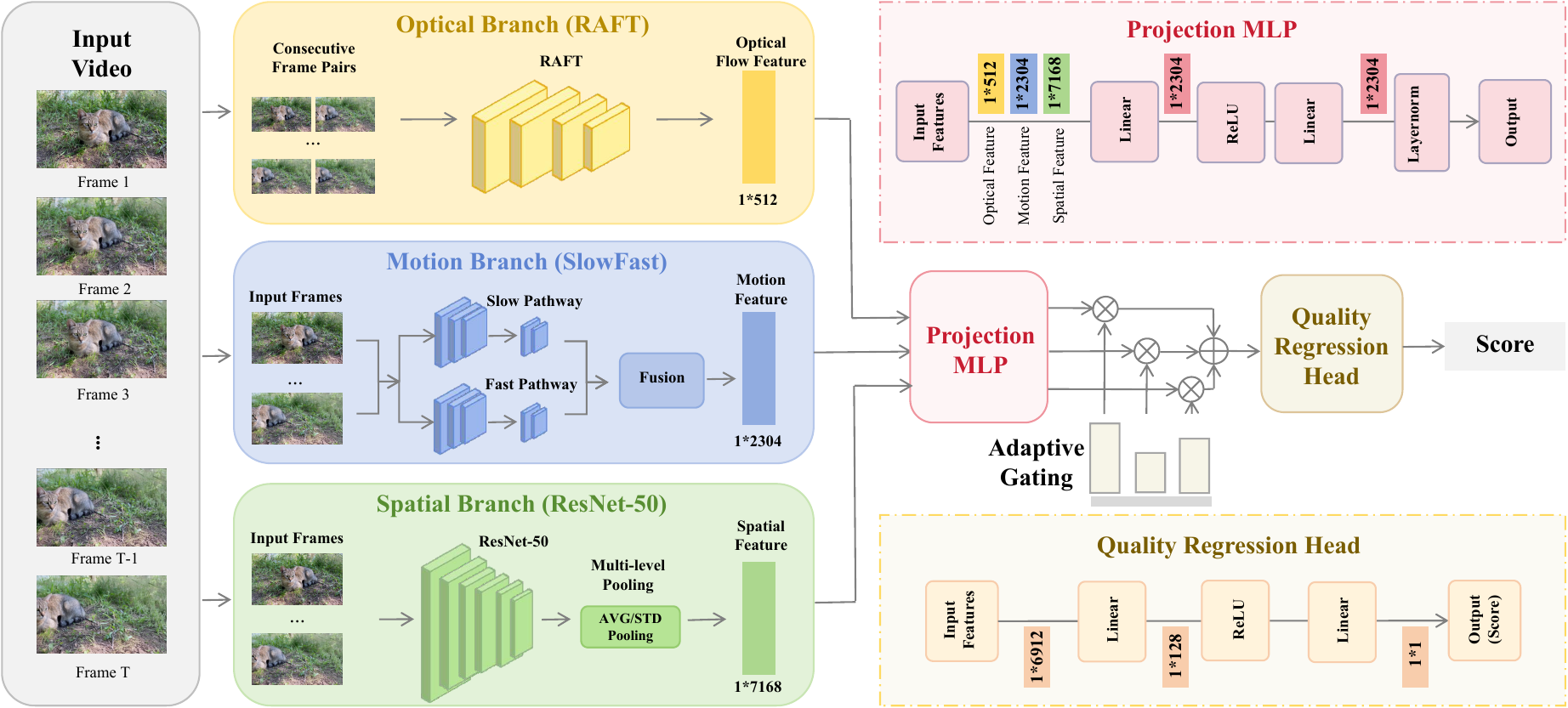}
    \caption{Overview of the proposed CWQA method.}
    \label{fig:overview}
    \vspace{-2mm}
\end{figure*}

\subsection{Subjective Study}
To obtain reliable human perceptual quality annotations, we conducted a
laboratory-based subjective study following ITU-R BT.500 \cite{itu500}.
The study involved 18 human subjects with a balanced gender distribution.
Videos were organized by content category and camera trajectory, and were
presented in random order to reduce method-ordering bias.
Each video was rated with an overall quality score from 1 to 5, with one
decimal place allowed.
Participants were allowed to replay the video before submitting their scores.
During rating, participants jointly considered \textbf{Viewpoint Consistency},
\textbf{Motion Coherence}, and \textbf{Content Preservation}.
These three aspects evaluate plausible and stable viewpoints, smooth motions
without jitter or flickering, and complete main subjects without unintended
changes or severe deformation, respectively.

\subsection{Data processing}
Following ITU-R BT.500 \cite{itu500}, we detected stimulus-level outliers
based on the mean, standard deviation, and Pearson kurtosis of each
stimulus's scores.
Subjects with excessive outlier counts were rejected.
The remaining scores were normalized subject-wise, and the MOS of each
stimulus was computed as the mean of its normalized valid ratings:

\begin{equation}
z_{ij} = \frac{r_{ij} - \mu_i}{\sigma_i}, \quad
z'_{ij} = \frac{100(z_{ij} + 3)}{6}, \quad
\mathrm{MOS}_j = \frac{1}{|\mathcal{V}_j|}
\sum_{i\in\mathcal{V}_j} z'_{ij},
\end{equation}

where $\mu_i$ and $\sigma_i$ are the mean and standard deviation of subject
$i$'s raw scores, and $\mathcal{V}_j$ denotes the valid ratings for
stimulus $j$.
In total, 18 subjects rated all 720 stimuli. One subject was excluded based on the quality control criteria, and the final MOS for each stimulus was computed from the remaining valid ratings.

\subsection{Data Analysis}
\label{sec:data-analysis}

To better illustrate CamWorldQA, we analyze the MOS distributions across
different dimensions, as shown in Table~\ref{tab:mos-statistics} and
Figure~\ref{fig:model-dist} and Figure~\ref{fig:overall-dist}.
\textbf{Overall}, the MOS distribution presents two peaks at high and low quality
levels and is largely concentrated within the range of 20 to 80.
This wide coverage allows CamWorldQA to differentiate clearly degraded
outputs from highly realistic ones and provides a solid basis for evaluating
quality assessment methods.
From the \textbf{method-wise} perspective, the six generation methods form two
distinct quality tiers: GEN3C, ReDirector, SierpinskiCam, and ReCamMaster
consistently achieve higher average MOS, while Diffusion as Shader and
TrajectoryCrafter fall clearly behind, leaving a marked gap between the
two groups.
From the \textbf{trajectory-wise} perspective, the six camera trajectories show only
modest quality differences.
Dolly-based movements are rated slightly higher and pedestal-down movements
slightly lower, yet no trajectory yields a systematic advantage.
From the \textbf{content-wise} perspective, the four content categories obtain
similar average MOS, suggesting that the observed quality characteristics
are not tied to any single content type and that our dataset generalizes
well across diverse video contents.

\section{Method}

\subsection{Overview}

As illustrated in Fig.~\ref{fig:overview}, CWQA takes a camera-controlled generated video as input
and predicts its perceptual quality score. The framework consists of three feature representation branches, an
adaptive feature fusion module, and a quality regression module.

Specifically, the spatial branch extracts appearance
features, while the temporal branch captures high-level video dynamics and the optical flow branch is designed to characterize motion
information. After mapping them into a common feature space,
the projected features are then
jointly fed into a branch-wise gating module, adaptively adjusting
the contribution of each representation. Finally, the weighted features
are concatenated to predict
the final video quality score.

\subsection{Feature Representation}

Camera-controlled generated videos exhibit quality degradations in
spatial appearance, temporal dynamics, and camera-induced motion.
Accordingly, we extract three complementary representations to adaptively characterize these quality variations.

\noindent\textbf{Spatial Branch.}
Camera-controlled generation may introduce appearance-related
degradations, such as local distortions and structural deformation. To characterize
these spatial quality variations, we extract multi-level appearance
features from sampled RGB frames~\cite{sun2022deep}. Features from different stages
are aggregated using average and standard deviation pooling to capture
global appearance information and local feature variations. The resulting
spatial representation is denoted as $F_s$.

\noindent\textbf{Temporal Branch.}
In addition to spatial distortions, camera-controlled videos may exhibit
temporal degradations, such as flickering and
unstable content across frames. To capture these temporal quality
variations, we extract high-level video dynamics using SlowFast
features~\cite{feichtenhofer2019slowfast} across sampled frames. The resulting temporal
representation is denoted as $F_t$.

\noindent\textbf{Optical Flow Branch.}
Camera manipulation introduces explicit motion patterns across consecutive
frames, making motion consistency an important factor in perceptual
quality. Generated videos may exhibit irregular camera-induced motion or unstable viewpoint transitions that are
not fully characterized by appearance and high-level temporal features.
To explicitly capture such motion variations, we extract optical flow between consecutive frames and encode it into motion representations~\cite{kou2023stablevqa}. The resulting optical flow
representation is denoted as $F_f$.

Since the three representations have different feature dimensions, we
employ separate projection modules to map them into a common feature
space:

\begin{equation}
\tilde{F}_s=P_s(F_s), \quad
\tilde{F}_t=P_t(F_t), \quad
\tilde{F}_f=P_f(F_f),
\end{equation}

where $P_s(\cdot)$, $P_t(\cdot)$, and $P_f(\cdot)$ denote the
corresponding feature projection modules. The projected representations
are subsequently used for adaptive feature fusion.
\subsection{Adaptive Branch-wise Feature Fusion}

Since the contributions of spatial, temporal, and optical flow
representations may vary across videos, we introduce adaptive branch-wise
fusion to dynamically weight the three representations. Given
$\tilde{F}_s$, $\tilde{F}_t$, and $\tilde{F}_f$, their concatenation is
fed into a lightweight gating network:

\begin{equation}
[g_s,g_t,g_f]
=
\sigma
\left(
G([\tilde{F}_s;\tilde{F}_t;\tilde{F}_f])
\right),
\end{equation}

where $G(\cdot)$ denotes the learnable gating network, $\sigma(\cdot)$
is the sigmoid activation function, and $g_s$, $g_t$, and $g_f$ denote
the weights assigned to the spatial, temporal, and optical flow
branches, respectively. The weights are applied as:

\begin{equation}
F_s^{w}=g_s\tilde{F}_s,\quad
F_t^{w}=g_t\tilde{F}_t,\quad
F_f^{w}=g_f\tilde{F}_f.
\end{equation}

The weighted representations are subsequently combined for quality
prediction.

\subsection{Quality Regression}

The weighted representations are concatenated to form the final
quality-aware representation:

\begin{equation}
F_q=[F_s^{w};F_t^{w};F_f^{w}].
\end{equation}

A quality regression head then predicts the perceptual quality score:

\begin{equation}
q_t=R(F_q),
\end{equation}

where $R(\cdot)$ denotes the quality regression function and $q_t$
represents the predicted quality score for the sampled frame position.
Predictions from all sampled positions are averaged to obtain the
video-level quality score~\cite{sun2022deep}:

\begin{equation}
Q=\frac{1}{T}\sum_{t=1}^{T}q_t,
\end{equation}

where $T$ denotes the number of sampled positions and $Q$ is the final
predicted perceptual quality score.

\section{Experiments}

\subsection{Experimental Setup}

\noindent\textbf{Dataset and Split.}
We conduct experiments on CamWorldQA, which contains 720
camera-controlled generated videos with subjective MOS annotations.
The videos are generated from 20 real-world source videos covering four
content categories using 6 generation methods and 6 representative
camera trajectories. The dataset is divided into training and testing
sets with a ratio of 8:2 using a fixed random seed. The split is
performed at the generated-video level, ensuring that each generated
video appears in only one subset.

\noindent\textbf{Training Details.}
CWQA is initialized from a SimpleVQA~\cite{sun2022deep}
checkpoint that has been
previously fine-tuned on the CamWorldQA training set. During training, the original feature extraction backbones
are frozen, while the spatial, temporal, and optical flow projection
modules, the adaptive branch-wise gating module, and the quality
regression head are optimized. We employ the Adam optimizer with the L1 ranking loss. The learning rate
is set to $1\times10^{-4}$ for the three feature projection modules and
the branch-wise gating module, and $1\times10^{-5}$ for the quality
regression head, with a weight decay of $1\times10^{-7}$. A StepLR
scheduler is adopted to decay the learning rate by a factor of 0.95
every two epochs. The model is trained for 20 epochs with a batch size
of 4. For data preprocessing, input frames are resized to 520 pixels and
randomly cropped to $448\times448$ during training, while center cropping
is used during testing. ImageNet normalization is applied to the RGB
frames.

\begin{table}[ht]
\centering
\caption{Overall performance comparison on the CamWorldQA dataset. $\diamondsuit$, $\clubsuit$, and $\heartsuit$ denote traditional VQA methods, learning-based VQA methods, and multimodal large language models, respectively. $^{\dagger}$ indicates models trained on the CamWorldQA dataset. The best and runner-up performances are bold and underlined, respectively.}
\label{tab:comparison}
\resizebox{0.9\columnwidth}{!}{
\begin{tabular}{l|ccc}
\toprule
Method & SRCC\,$\uparrow$ & PLCC\,$\uparrow$ & KRCC\,$\uparrow$ \\
\midrule

$\diamondsuit$ RAPIQUE~\cite{tu2021rapique}
& \underline{0.6592} & \underline{0.6545} & \underline{0.4722} \\

$\diamondsuit$ VIDEVAL~\cite{tu2021ugc}
& 0.3491 & 0.3677 & 0.2386 \\

$\clubsuit$ SimpleVQA~\cite{sun2022deep}
& 0.1816 & 0.1906 & 0.1212 \\

$\clubsuit$ FastVQA~\cite{wu2022fast}
& 0.4662 & 0.4615 & 0.3112 \\

$\clubsuit$ DOVER~\cite{wu2023exploring}
& 0.2098 & 0.2590 & 0.1428 \\

$\heartsuit$ Qwen3-VL (4B)~\cite{bai2025qwen3}
& 0.2342 & 0.2526 & 0.1852 \\

$\heartsuit$ VideoLLaMA3 (7B)~\cite{zhang2025videollama}
& 0.3460 & 0.3843 & 0.2710 \\

$\heartsuit$ Qwen3-VL (8B)~\cite{bai2025qwen3}
& 0.2640 & 0.2819 & 0.2139 \\

$\heartsuit$ InternVL3.5 (8B)~\cite{wang2025internvl3}
& 0.1147 & 0.1557 & 0.0935 \\


\midrule

$\clubsuit$ SimpleVQA~\cite{sun2022deep} $^{\dagger}$
& 0.6263 & 0.6260 & 0.4478 \\

$\clubsuit$ FastVQA~\cite{wu2022fast} $^{\dagger}$
& 0.5006 & 0.4948 & 0.3405 \\

$\clubsuit$ DOVER~\cite{wu2023exploring} $^{\dagger}$
& 0.4236 & 0.4265 & 0.2925 \\

\rowcolor[gray]{.92}
\textbf{CWQA (Ours)} $^{\dagger}$
& \textbf{0.7804}
& \textbf{0.7848}
& \textbf{0.5739} \\

\bottomrule
\end{tabular}
}
\vspace{-3mm}
\end{table}

\begin{table*}[t]
\centering
\caption{
Category-wise performance comparison on the CamWorldQA dataset.
$\diamondsuit$, $\clubsuit$, and $\heartsuit$ denote traditional VQA methods, learning-based VQA methods, and multimodal large language models, respectively.
$^{\dagger}$ indicates models trained on the CamWorldQA dataset. The best and runner-up performances are bold and underlined, respectively.
}
\vspace{-0.5mm}
\label{tab:category}

\resizebox{0.9\textwidth}{!}{
\begin{tabular}{l|ccc|ccc|ccc|ccc}
\toprule
& \multicolumn{3}{c|}{Animals}
& \multicolumn{3}{c|}{Human}
& \multicolumn{3}{c|}{Objects}
& \multicolumn{3}{c}{Scene} \\
\cmidrule(lr){2-4} \cmidrule(lr){5-7} \cmidrule(lr){8-10} \cmidrule(lr){11-13}
Method
& SRCC\,$\uparrow$ & PLCC\,$\uparrow$ & KRCC\,$\uparrow$
& SRCC\,$\uparrow$ & PLCC\,$\uparrow$ & KRCC\,$\uparrow$
& SRCC\,$\uparrow$ & PLCC\,$\uparrow$ & KRCC\,$\uparrow$
& SRCC\,$\uparrow$ & PLCC\,$\uparrow$ & KRCC\,$\uparrow$ \\
\midrule

$\diamondsuit$ RAPIQUE~\cite{tu2021rapique}
& 0.7051 & \underline{0.7430} & 0.5352
& \underline{0.7096} & \textbf{0.7780} & \underline{0.4970}
& 0.5003 & 0.7114 & 0.3492
& 0.7156 & 0.6880 & \textbf{0.5116} \\

$\diamondsuit$ VIDEVAL~\cite{tu2021ugc}
& 0.2948 & 0.2662 & 0.2135
& 0.3206 & 0.4011 & 0.2182
& 0.1970 & 0.5798 & 0.1429
& 0.5184 & 0.5230 & 0.3693 \\

$\clubsuit$ DOVER~\cite{wu2023exploring}
& 0.1844 & 0.2516 & 0.1380
& 0.0392 & 0.0237 & 0.0242
& 0.3689 & 0.6359 & 0.2751
& 0.4226 & 0.3792 & 0.3063 \\

$\clubsuit$ SimpleVQA~\cite{sun2022deep}
& 0.2836 & 0.3574 & 0.2164
& 0.0460 & 0.3703 & 0.0364
& 0.0317 & 0.0480 & 0.0159
& 0.3968 & 0.4080 & 0.3171 \\

$\clubsuit$ FastVQA~\cite{wu2022fast}
& 0.3117 & 0.3893 & 0.2006
& 0.3138 & 0.3723 & 0.1855
& 0.4165 & 0.5740 & 0.2698
& 0.7097 & \underline{0.8089} & 0.4957 \\

$\heartsuit$ Qwen3-VL (4B)~\cite{bai2025qwen3}
& 0.2137 & 0.2899 & 0.1792
& 0.1389 & 0.2927 & 0.1104
& 0.1648 & 0.2728 & 0.1440
& 0.4262 & 0.3715 & 0.3484 \\

$\heartsuit$ VideoLLaMA3 (7B)~\cite{zhang2025videollama}
& 0.4475 & 0.4431 & 0.3659
& 0.2303 & 0.3479 & 0.1909
& 0.3232 & 0.3316 & 0.2833
& 0.4964 & 0.6185 & 0.4028 \\

$\heartsuit$ Qwen3-VL (8B)~\cite{bai2025qwen3}
& 0.2090 & 0.1969 & 0.1729
& 0.4578 & 0.4567 & 0.3796
& 0.1514 & 0.1197 & 0.1257
& 0.4162 & 0.4107 & 0.3493 \\

$\heartsuit$ InternVL3.5 (8B)~\cite{wang2025internvl3}
& 0.0796 & 0.1021 & 0.0659
& 0.5159 & 0.5266 & 0.4277
& 0.0210 & 0.2890 & 0.0244
& 0.1509 & 0.1139 & 0.1246 \\


\midrule

$\clubsuit$ DOVER~\cite{wu2023exploring} $^{\dagger}$
& 0.5404 & 0.6007 & 0.3855
& 0.5561 & 0.5975 & 0.3831
& 0.4444 & 0.4655 & 0.3492
& 0.4187 & 0.2336 & 0.2812 \\

$\clubsuit$ SimpleVQA~\cite{sun2022deep} $^{\dagger}$
& \underline{0.7426} & 0.7318 & \underline{0.5352}
& 0.6330 & 0.7196 & 0.4525
& \underline{0.7061} & 0.7220 & \underline{0.5026}
& 0.5578 & 0.5500 & 0.3809 \\

$\clubsuit$ FastVQA~\cite{wu2022fast} $^{\dagger}$
& 0.3812 & 0.4686 & 0.2875
& 0.2700 & 0.3868 & 0.1756
& 0.5012 & \underline{0.7382} & 0.3364
& \underline{0.7189} & \textbf{0.8207} & 0.5039 \\

\rowcolor[gray]{.92}
\textbf{CWQA (Ours)} $^{\dagger}$
& \textbf{0.8601} & \textbf{0.8490} & \textbf{0.6776}
& \textbf{0.7115} & \underline{0.7345} & \textbf{0.5131}
& \textbf{0.8380} & \textbf{0.8828} & \textbf{0.6402}
& \textbf{0.7239} & 0.7659 & \underline{0.5085} \\

\bottomrule
\end{tabular}
}
\end{table*}

\noindent\textbf{Compared Methods.}
We compare our method with representative quality assessment approaches
under different evaluation settings. Conventional no-reference VQA
models, including DOVER~\cite{wu2023exploring},
SimpleVQA~\cite{sun2022deep}, and
FastVQA~\cite{wu2022fast}, are evaluated under both zero-shot and
fine-tuned settings. We additionally include VIDEVAL~\cite{tu2021ugc} and RAPIQUE~\cite{tu2021rapique} for zero-shot evaluation . We further include recent video multimodal large
language models, including the Qwen3 series~\cite{bai2025qwen3},
InternVL3.5 series~\cite{wang2025internvl3},
and VideoLLaMA3~\cite{zhang2025videollama}, to provide
a broader comparison with general-purpose video understanding models.
The zero-shot setting evaluates direct transferability to camera-controlled
generated videos, whereas the fine-tuned setting evaluates adaptation
using the subjective annotations of CamWorldQA.

\noindent\textbf{Evaluation Metrics.}
Following previous VQA studies, we employ three commonly used
correlation metrics: Pearson Linear Correlation Coefficient (PLCC),
Spearman Rank Correlation Coefficient (SRCC), and Kendall Rank
Correlation Coefficient (KRCC). PLCC measures the linear correlation
between predicted quality scores and subjective ratings, while SRCC and
KRCC measure ranking consistency. Before calculating PLCC, nonlinear
logistic mapping is applied following the adopted VQA evaluation
protocol.

\subsection{Ablation Study}

We conduct ablation experiments to investigate the contribution of each
feature branch and the adaptive fusion strategy. All variants are trained
and evaluated under the same experimental settings, with the results
reported in Table~\ref{tab:ablation}.

\begin{table}[ht]
\centering
\caption{Ablation study of CWQA on CamWorldQA.}
\vspace{-1mm}
\label{tab:ablation}
\resizebox{0.8\columnwidth}{!}{
\begin{tabular}{l|ccc}
\toprule
Method & SRCC\,$\uparrow$ & PLCC\,$\uparrow$ & KRCC\,$\uparrow$ \\
\midrule
w/o Spatial Branch
& 0.7572 & 0.7553 & 0.5712 \\

w/o Temporal Branch
& 0.7248 & 0.7099 & 0.5308 \\

w/o Optical Flow
& 0.6263 & 0.6260 & 0.4478 \\

w/o Adaptive Gating
& 0.7315 & 0.6636 & 0.5374 \\

\rowcolor[gray]{.92}
\textbf{Full CWQA (Ours)}
& \textbf{0.7804}
& \textbf{0.7848}
& \textbf{0.5739} \\
\bottomrule
\end{tabular}
}
\vspace{-3mm}
\end{table}

As shown in Table~\ref{tab:ablation}, removing the spatial branch leads
to a moderate performance degradation, indicating that multi-level
appearance features provide useful information for capturing spatial
distortions and structural deformation. A more noticeable degradation is
observed without the temporal branch, demonstrating the importance of
high-level temporal features in characterizing temporal variations and
cross-frame instability. Among the three feature branches, removing the
optical flow branch causes the most significant performance drop,
highlighting the role of explicit motion information in capturing
irregular camera-induced motion and unstable viewpoint transitions. The adaptive gating module further improves the overall performance, indicating that dynamically weighting the three representations is more effective than directly combining them. The complete CWQA achieves the best performance, demonstrating the complementarity of the three feature branches and the effectiveness of adaptive feature fusion.

\subsection{Results Analysis}

As shown in Table~\ref{tab:comparison}, existing VQA and multimodal models
show limited performance when directly applied to CamWorldQA, indicating
a domain gap between conventional quality representations and
camera-controlled generated videos. Fine-tuning on CamWorldQA generally
improves the performance of conventional VQA methods, demonstrating the
benefit of task-specific quality annotations. Among the compared methods,
CWQA achieves the best performance across all three correlation metrics
and clearly outperforms the fine-tuned baselines. This improvement
demonstrates the effectiveness of jointly modeling spatial, temporal,
and optical flow information for perceptual quality assessment of
camera-controlled generated videos.

To examine the robustness of different methods across diverse video
contents, we further report category-wise results in
Table~\ref{tab:category}. The compared methods exhibit noticeable
performance variations across different content categories. In
particular, several zero-shot VQA and multimodal models perform well on
specific categories but show limited generalization to others,
suggesting that their quality representations do not transfer uniformly
to camera-controlled generated videos. Fine-tuning on CamWorldQA
generally improves the performance of conventional VQA models,
highlighting the importance of task-specific adaptation. CWQA achieves consistently strong correlations with subjective
scores across the four categories. It achieves the best SRCC across all four categories and the best PLCC on animals and objects, while remaining competitive with the strongest baselines on human and scene videos. These results indicate that jointly
modeling spatial, temporal, and optical flow representations provides a
more robust quality representation across different types of generated
content.

\begin{figure}[t]
    \centering
    \includegraphics[width=\columnwidth]{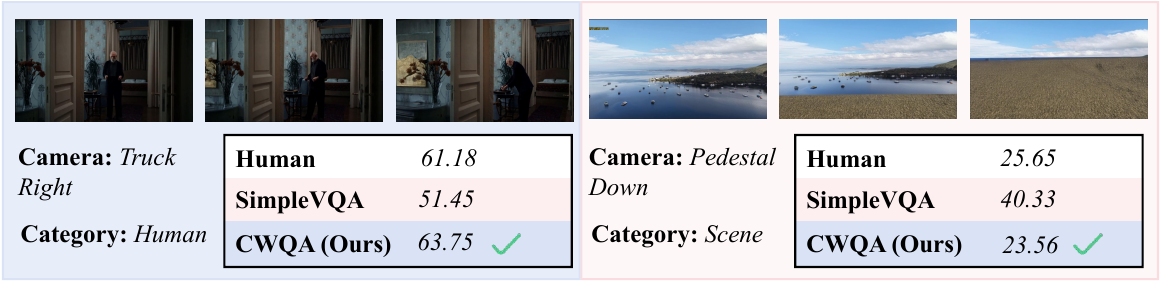}
    \caption{Qualitative results of the proposed CWQA method.}
    \label{fig:resultsdemo}
    \vspace{-5mm}
\end{figure}
As illustrated in Fig.~\ref{fig:resultsdemo}, CWQA produces quality predictions that are more consistent with human MOS than the compared methods across representative cases with different perceptual degradations. This qualitative comparison further demonstrates the effectiveness of CWQA in capturing quality variations in camera-controlled generated videos.

\section{Conclusion}

In this work, we present \textbf{CamWorldQA}, the first benchmark for perceptual quality assessment of camera-controlled world video generation. CamWorldQA contains 720 generated videos produced by 6 representative generation methods from 20 diverse source videos under 6 camera trajectories, where each video is annotated with a human-rated perceptual quality score through subjective experiments. Furthermore, we propose \textbf{CWQA}, a no-reference quality assessment network with three complementary branches that extract spatial features, temporal motion features and optical flow features to jointly predict quality scores. Extensive experiments demonstrate that CWQA achieves superior performance over existing quality assessment methods on the CamWorldQA dataset. We hope CamWorldQA can facilitate future research on the evaluation and improvement of camera-controlled video generation.




\appendix
\clearpage
\renewcommand{\thetable}{\Alph{table}}
\renewcommand{\thefigure}{\Alph{figure}}
\setcounter{table}{0}
\setcounter{figure}{0}

\twocolumn[
\begin{center}
    \huge\bfseries Supplementary Material
    \vspace{2mm}
\end{center}
]

\section{Trajectory Control}

We define a canonical camera-motion space to unify different trajectory
settings, as shown in Table~\ref{tab:canonical-camera-parameters}.
Each trajectory is represented by three camera parameters: 
$t_x$, $t_y$, and $t_z$.

The parameters $t_x$, $t_y$, and $t_z$ denote the camera's translational displacements along the three principal axes of the canonical coordinate system: 
$t_x$ for horizontal shift (positive = right, negative = left), 
$t_y$ for vertical shift (positive = up, negative = down), 
and $t_z$ for depth shift along the viewing direction (positive = moving away from the scene, negative = moving toward the scene). 
These values are expressed in normalized units relative to the scene scale, ensuring consistency across different datasets and rendering settings. The chosen magnitudes (e.g., $\pm0.25$ for lateral/vertical, $\pm0.30$ for depth) are empirically selected to produce noticeable but non-extreme motion effects, facilitating robust learning of motion-aware representations.

\begin{table}[t]
\centering
\caption{Canonical camera motions and their corresponding camera parameters.}
\label{tab:canonical-camera-parameters}
\begin{tabular}{lccc}
\toprule
\textbf{Canonical Motion} &
\textbf{$t_x$} & \textbf{$t_y$} & \textbf{$t_z$} \\
\midrule
Truck Left    & $-0.25$ & $0$ & $0$ \\
Truck Right   & $+0.25$ & $0$ & $0$ \\
Pedestal Up   & $0$ & $+0.25$ & $0$ \\
Pedestal Down & $0$ & $-0.25$ & $0$ \\
Dolly In      & $0$ & $0$ & $-0.30$ \\
Dolly Out     & $0$ & $0$ & $+0.30$ \\
\bottomrule
\end{tabular}
\end{table}

\section{Camera-controlled Video Generation Methods}

\noindent\textbf{ReCamMaster} \cite{bai2025recammaster} : A camera-controlled generative rendering
method that synthesizes novel viewpoints of a single video under
user-specified camera trajectories. It learns a camera-conditioned
diffusion model built on a Flux DiT backbone with a CogVAE, injecting
camera parameters as conditioning to control viewpoint changes. It
employs 3D-consistent rendering with view-dependent appearance modeling,
trained on 81-frame 1280$\times$1280 15 fps videos. It supports diverse
camera trajectories (e.g., dolly, truck, and pedestal movements), making
it suitable for film production and interactive applications.

\noindent\textbf{TrajectoryCrafter} \cite{yu2025trajectorycrafter} : Redirects the camera trajectory of
monocular videos through diffusion-based re-rendering. It first estimates
dense depth with DepthCrafter, warps the scene along the target
trajectory, and fills disoccluded regions via a spatio-temporal U-Net for
temporal consistency. It generates 49-frame clips at 10 fps, using the
depth prior as an explicit camera-control signal. Its depth-guided
re-rendering enables seamless camera retargeting on casual monocular
videos.

\noindent\textbf{ReDirector} \cite{park2026redirector} : Creates any-length video retakes with a
rotary camera encoding mechanism. It embeds camera motion as rotary
positional embeddings into a Wan 2.1-Fun 1.3B backbone, enabling dense
camera control without explicit depth estimation or 3D reconstruction,
and supports arbitrary-length retakes at 832$\times$480 resolution. By
separating camera control from content, it allows flexible retakes of
existing videos with any desired camera motion and duration.

\noindent\textbf{GEN3C} \cite{ren2025gen3c} : A 3D-informed world-consistent video generation
method with precise camera control. It maintains an explicit 3D cache of
the scene, renders it from the target viewpoint, and feeds the renderings
into an NVIDIA Cosmos autoregressive model with a 7B diffusion decoder.
It supports long 121-frame generations at 24 fps, ensuring geometric
consistency under large camera motions. It operates on a single image or
video seed and generates very long sequences autoregressively.

\noindent\textbf{Diffusion as Shader} \cite{gu2025diffusion} : Treats the video diffusion model
as a shader that re-renders an input 3D mesh under novel camera motion.
Built on CogVideoX-5B, it re-projects rendered geometry back into the
diffusion pipeline, providing interpretable 3D-aware control over camera
pan, zoom, and object motion, while generating 49 frames at 8 fps.
Leveraging explicit mesh rendering and point tracking, it offers precise,
user-controllable 3D-aware motion editing.

\noindent\textbf{SierpinskiCam} \cite{wizadwongsa2026sierpinskicam} : Camera-controlled video retaking using
Sierpinski-triangle pattern cues. It encodes camera trajectories as
Sierpinski pattern signals injected into a Wan 2.1 14B diffusion model,
enabling arbitrary camera paths for retaking at 832$\times$480 and 81
frames, without requiring per-frame camera parameters at inference. Its
pattern-based camera cues support 14 predefined camera paths and avoid
explicit camera parameter estimation.

The overview of six camera-controlled video generation models is shown in Table \ref{tab:models}

\begin{table}[t]
\centering
\small
\setlength{\tabcolsep}{4pt}
\caption{Overview of the six camera-controlled video generation methods.}
\label{tab:models}
\begin{tabular}{lcccc}
\toprule
Method & Year & Resolution & Frames (FPS) & Type \\
\midrule
ReCamMaster   & 2025 & 1280$\times$1280\ddag & 81\ddag/15\ddag & Diff.\ (3D) \\
TrajectoryCrafter & 2025 & Inherited$^*$ & 49\dag/10\dag & Diffusion \\
ReDirector    & 2025 & 832$\times$480\dag  & 81\dag/30\dag  & Diffusion \\
GEN3C         & 2025 & 704$\times$1280\ddag & 121\ddag/24\dag & 3D world \\
\shortstack{Diffusion as Shader} & 2025 & 720$\times$480\ddag & 49$^*/8\dag$ & 3D-aware \\
SierpinskiCam & 2026 & 832$\times$480\dag & 81\dag/12\dag & Diffusion \\
\bottomrule
\multicolumn{5}{l}{\footnotesize $\dag$ default; $^*$ maximum; $\ddag$ example/training-recommended.}
\end{tabular}
\end{table}

\section{Details of the Subjective Study GUI}

\begin{figure*}[t!]             
    \centering
    \includegraphics[
        width=0.95\textwidth,      
        height=0.42\textheight,   
        keepaspectratio
    ]{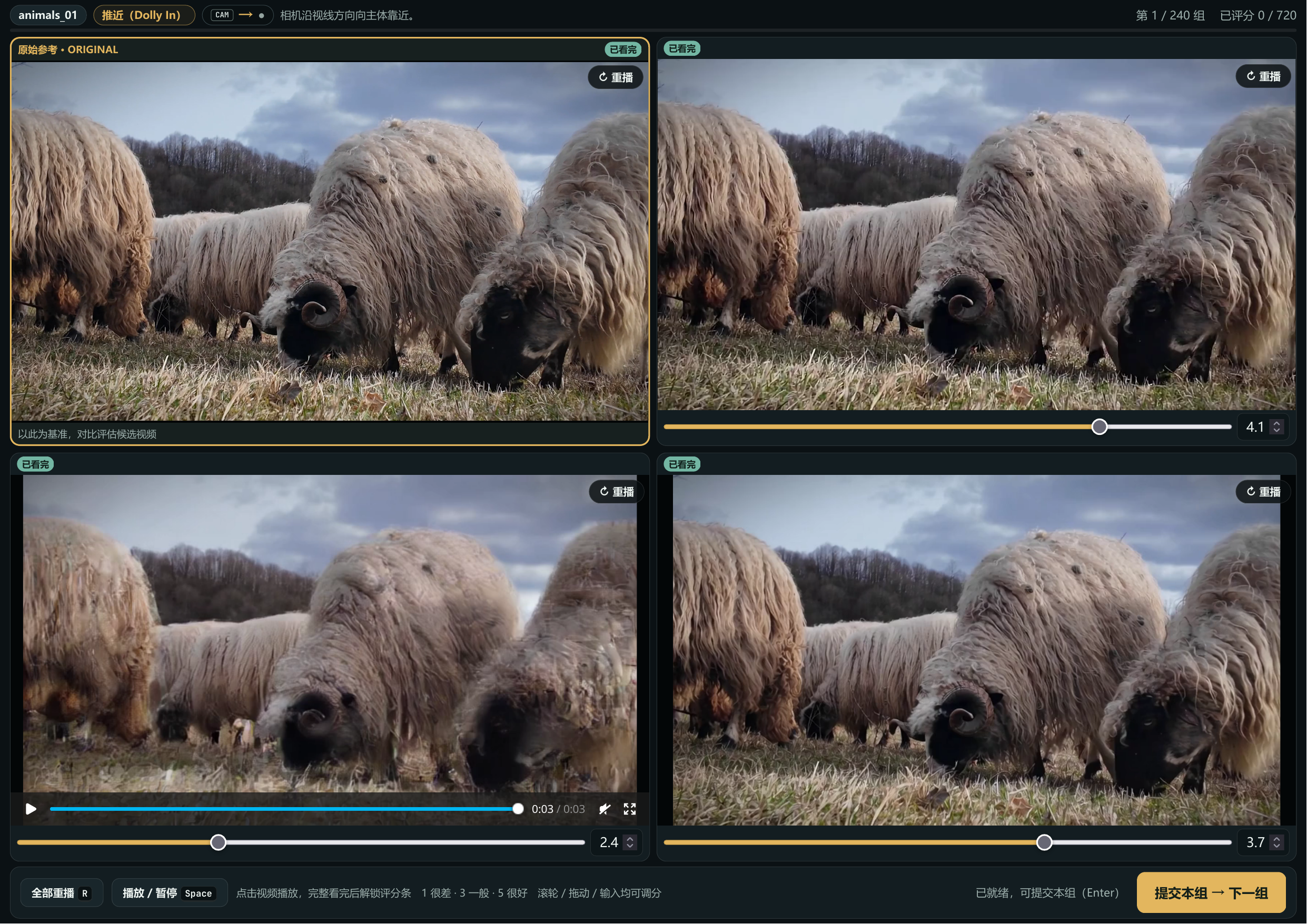}
    \caption{Illustration of the GUI used in the subjective study}
    \label{fig:gui}
\end{figure*}

We developed a browser-based GUI for perceptual quality assessment of camera-controlled video generation as shown in Figure \ref{fig:gui}. After entering a user ID, the interface starts or resumes the participant's progress with an individually randomized playback order. The rating screen shows the sample ID, trajectory name, a camera-motion diagram, and a progress bar. Videos are displayed in a card grid: a highlighted "Original Reference" on the left serves as the baseline, while three blind-labeled candidate videos are arranged on the right and bottom, with method names hidden and order randomized. Each candidate is rated via a 1.0–5.0 (0.1-step) slider, enabled only after the participant has fully watched the reference and all candidates. The bottom toolbar offers replay-all, play/pause, and submit shortcuts; the results are saved to an Excel workbook, and a completion screen shows the result file path.

\section{Demos for Videos of Different Mean Opinion Scores}
\begin{figure*}[htbp]
    \centering
    \vspace{1em}
    \includegraphics[width=0.8\linewidth]{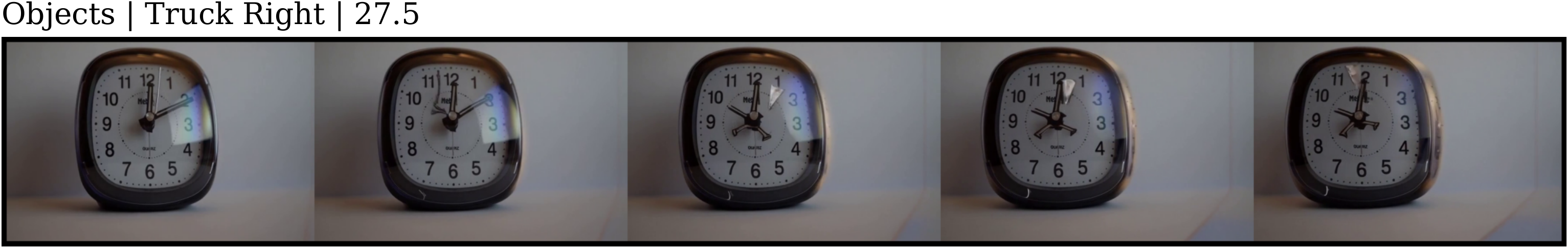}
    
    \vspace{1em}
    \includegraphics[width=0.8\linewidth]{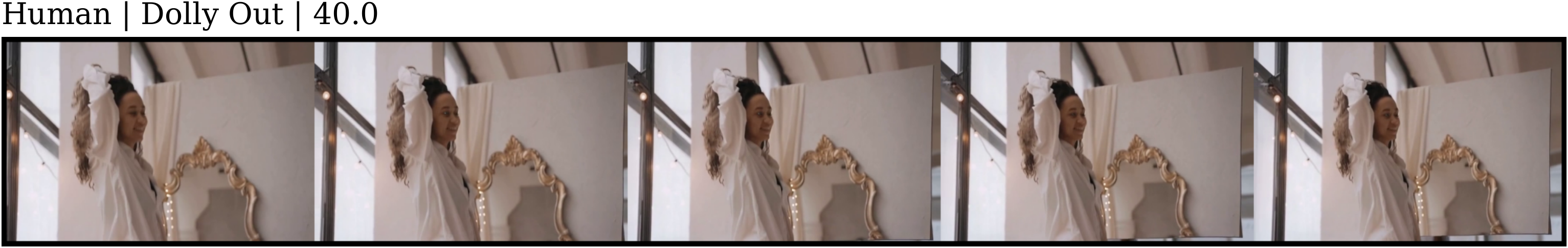}
    
    \vspace{1em}
    \includegraphics[width=0.8\linewidth]{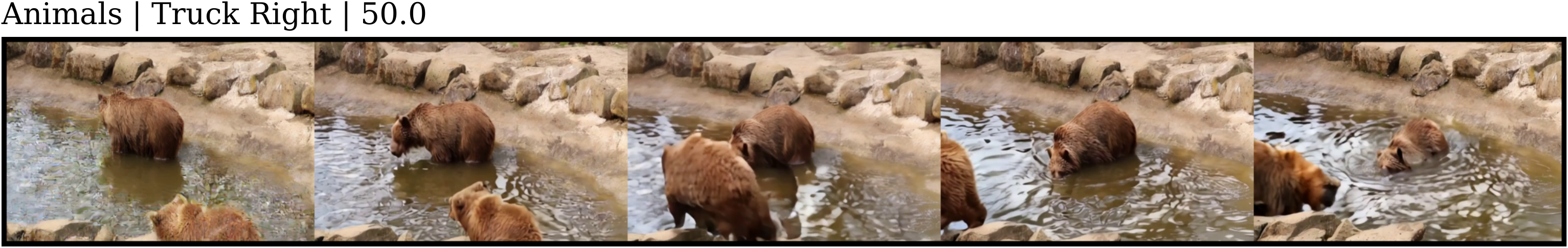}
    
    \vspace{1em}
    \includegraphics[width=0.8\linewidth]{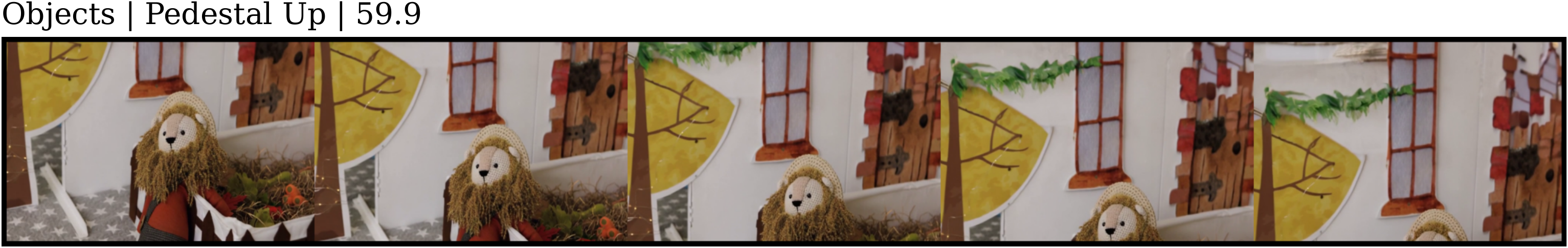}
    
    \vspace{1em}
    \includegraphics[width=0.8\linewidth]{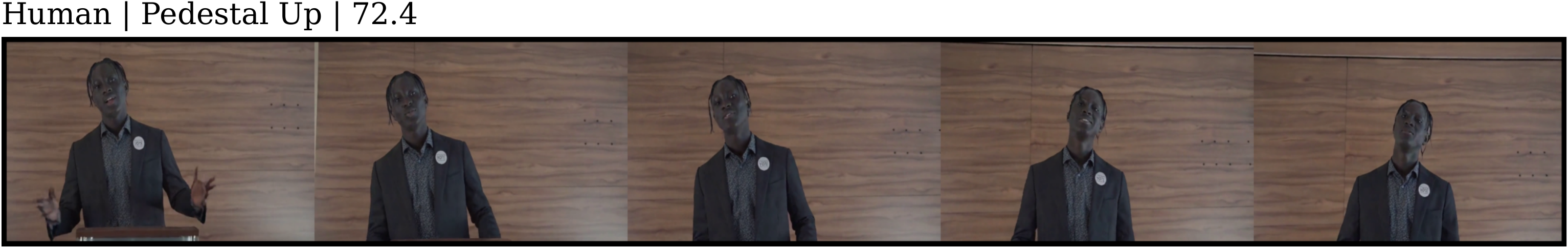}
    
    \caption{Demo videos of different mean opinion scores. For each video, its content category, camera trajectory, and MOS are annotated above the frames.} 
    \label{fig:demos}
\end{figure*}

To qualitatively illustrate how perceptual quality varies across different levels, we select five representative generated videos ranging from low to high MOS shown in Figure \ref{fig:demos}. \textbf{(1)} The lowest-quality example depicts a ticking clock, whose distorted clock hands appear geometrically implausible under the camera motion, indicating severe deformation of the moving object and a violation of physical realism. \textbf{ (2)} The second example shows a dancer; however, in the generated video the person becomes static and the wall is truncated, reflecting a failure to preserve motion dynamics and scene completeness. \textbf{(3)} The third example contains two brown bears playing in a stream, which is semantically plausible and natural in motion but blurry, noticeably lowering the perceived clarity. \textbf{(4)} The fourth example is a toy doll, overall clear and consistent with the real world but with slight blur and noise, representing only a mild quality degradation. \textbf{(5)} The best example shows a person giving a speech, where both the human motion and the surrounding environment are faithfully reproduced, appearing clear and vivid and yielding the highest perceptual quality among the five examples. These observations indicate that the perceptual quality of camera-controlled generated videos is jointly determined by geometric fidelity, motion plausibility, content preservation, and visual clarity, and the diverse artifacts across these examples highlight the limitations of conventional quality assessment models in capturing generation-specific distortions.

\bibliographystyle{ACM-Reference-Format}
\bibliography{acmart}

\end{document}